\documentclass[letterpaper, 10 pt, conference]{ieeeconf}
\IEEEoverridecommandlockouts
\usepackage{graphicx} 
\usepackage{amsmath} 
\usepackage{cite}

\usepackage{fancyhdr}
 
\graphicspath{{figures/}}

\title{\LARGE \bf
ATIS + SpiNNaker: a Fully Event-based Visual Tracking Demonstration
}

\author{Arren Glover, \textit{Member, IEEE}, Alan B. Stokes, Steve Furber, \textit{Fellow, IEEE} and Chiara Bartolozzi, \textit{Member, IEEE}%
\thanks{A.Glover, and C.Bartolozzi are with the iCub Facility, Istituto Italiano di Tecnologia, Italy. {\tt\small \{arren.glover, chiara.bartolozzi\}@iit.it}. A. B. Stokes and S. Furber are with the School of Computer Science, University of Manchester, United Kingdom. {\tt\small \{alan.stokes, steve.furber\}@manchester.ac.uk}%
}}

\begin{document}

\maketitle
\thispagestyle{fancy}

\begin{abstract}    
 
The Asynchronous Time-based Image Sensor (ATIS) and the Spiking Neural Network Architecture (SpiNNaker) are both neuromorphic technologies that ``unconventionally'' use binary spikes to represent information. The ATIS produces spikes to represent the change in light falling on the sensor, and the SpiNNaker is a massively parallel computing platform that asynchronously sends spikes between cores for processing. In this demonstration we show these two hardware used together to perform a visual tracking task. We aim to show the hardware and software architecture that integrates the ATIS and SpiNNaker together in a robot middle-ware that makes processing agnostic to the platform (CPU or SpiNNaker). We also aim to describe the algorithm, why it is suitable for the ``unconventional'' sensor and processing platform including the advantages as well as challenges faced.

\end{abstract}

\section{EVENT-BASED TRACKING}

Event-based cameras propose a change in the paradigm used for visual sensing and processing that is attracting increasing interest in the robotics and computer vision communities. The Asynchronous Time-based Image Sensor (ATIS) produces an asynchronous signal encoding only the change in light, typically caused by motion, detected in the analogue circuitry of the sensor itself~\cite{Posch2008}. The signal is compressed, has a megahertz pixel update frequency, and the sensor operates in a much higher dynamic range when compared to traditional CMOS image sensors; while still maintaining, at minimum, the same visual information~\cite{Kim2014, Iacono2018}.

The Spiking Neural Network Architecture (SpiNNaker) is a massively extendable parallel processing platform~\cite{Furber2014} designed to simulate extremely large asynchronous processing algorithms (including, but not limited to, Artificial Spiking Neural Networks) in hardware, in real-time. A single board (of which multiple can be adjoined) has over 800 cores with networking fabric to quickly send spikes (or other small packets of information) to each other.

Both these technologies have seen an increase in interest over the last years, but are still ``unconventional'' with regards to the main-stream robotics community. We demonstrate our hardware/software architecture that integrates both technologies into a robotics middle-ware that allows plug-and-play of different sensors and devices. We show a live demonstration of the system performing a visual tracking task, which takes advantage of the event-cameras inherent response to motion, and the fact that the target cannot `jump' between frames, as the signal is continuous in space and time. The SpiNNaker runs a particle filter~\cite{Glover2017a}, in which the particle hypotheses are distributed amongst the cores, such that there is only a minor impact on algorithm compute rate when the number of particles is increased.

\begin{figure}
    \centering
    \includegraphics[width=0.8\linewidth, angle=180]{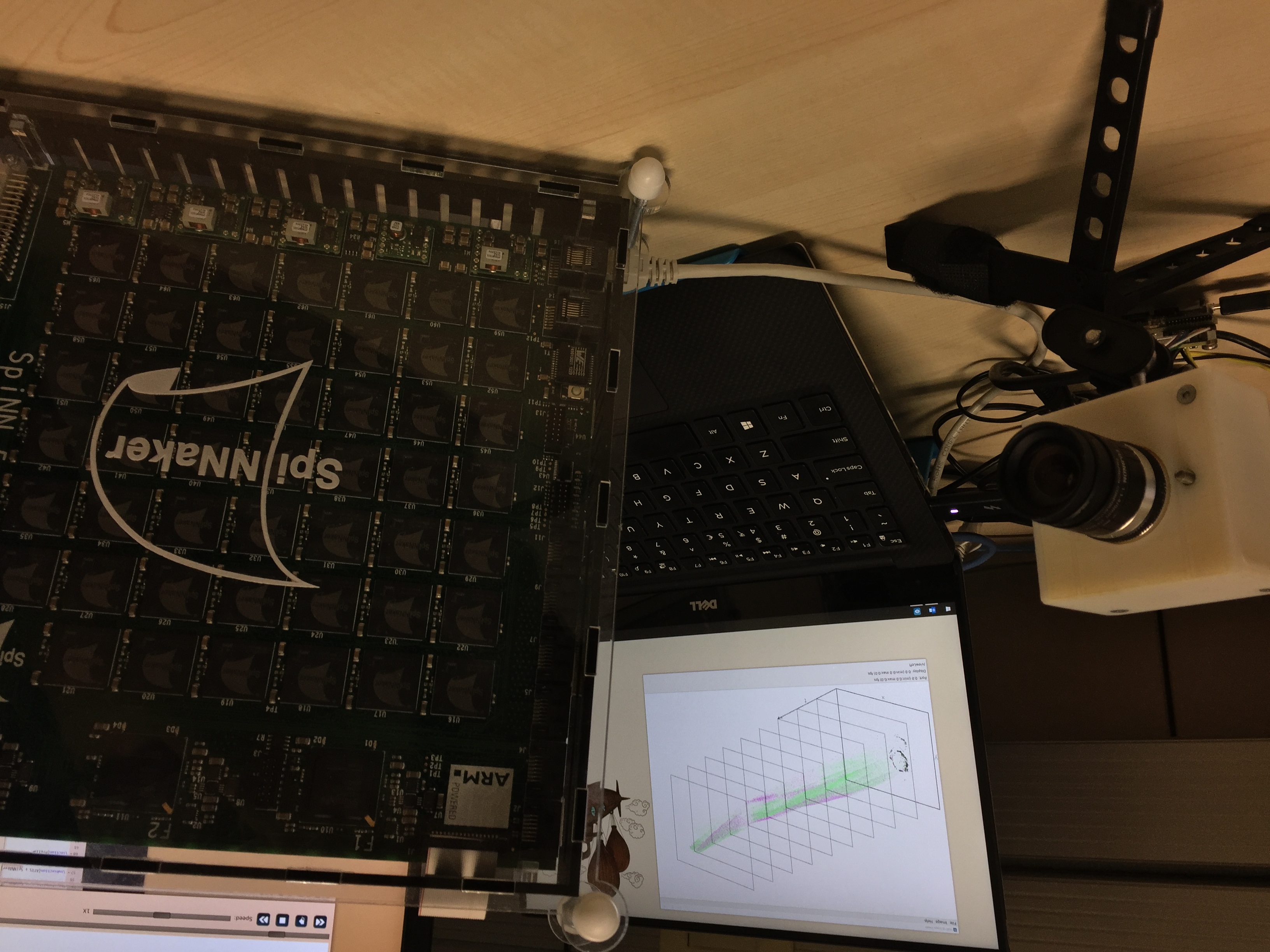}
    \caption{The ATIS event-camera and SpiNNaker computational platform used to perform a visual tracking task, with the result visualised on the laptop.}
    \label{fig:physical_stem}
\end{figure}

\subsection{Visual Tracking with a Particle Filter}
A particle filter estimates a system state (e.g. the position and size of a target) by maintaining a discrete set of hypothesis states and assigning a probabilistic weight to each hypothesis under a recursive Bayesian strategy. The Bayesian filter performs a feed-forward state update (based on modelled dynamics) with added noise, and then recalculates the probability based on the sensor values (or observations). If the expected observation from the updated state and the actual observation correspond, the hypothesis achieves a high probability. In the particle filter, each hypothesis is \textit{re-sampled} such that there are more particles representing highly probable states, and less-likely states are forgotten.

The visual tracking proposed in~\cite{Glover2017a} introduced the following innovations to adapt a particle filter to process the asynchronous stream of events produced by the ATIS camera:
\begin{itemize}
    \item Event-driven filter update, such that the filter is asynchronous and occurs more often when the target is moving faster.
    \item Algorithm for estimating the likelihood of a circle within the event-stream complying with the Markov assumption.
    \item Incremental calculation of the likelihood based on the temporal order of events to maximise the computed value, under an unknown temporal window.
    \item Proposal and validation of a constant-position motion model that is possible only as event-cameras have a continuous trajectory, which cannot `jump' between (non-existent) frames.
\end{itemize}
The algorithm was run on the iCub humanoid robot and demonstrated accurate tracking of a target circle, reaching $>2$ kHz when needed. It was also robust to false detections that can be caused by the background clutter produced as the robot moved the camera itself to follow the target with its gaze.

A particle filter typically has to compromise between accuracy and update-rate based on the number of particles. More particles result in more hypotheses, but a higher processing requirement. As the majority of the computation required for each particle can be performed independently from the others, it is therefore suitable for processing in parallel. If computed in parallel, the number of particles can be scaled without impact on the overall processing requirements. The SpiNNaker platform is designed for both asynchronous spiking input and for parallel processing, and therefore is a suitable choice for hardware acceleration, considering the spiking output of the ATIS. Each core of the SpiNNaker runs any loaded executable, and is not limited to models of spiking neurons. For this application an executable which performs the update steps of a single particle is loaded onto any number of cores of the SpiNNaker platform.

\begin{figure}
    \centering
    \includegraphics[width=\linewidth]{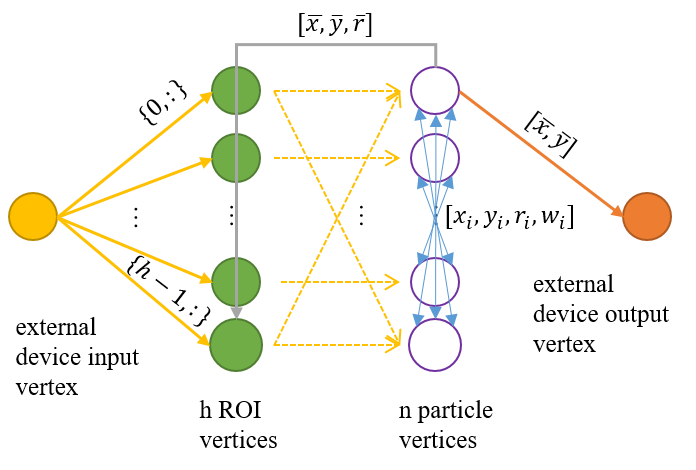}
    \caption{The network implemented on the SpiNNaker involves an input vertex which receives all events from the \textit{Xilinx Zynq}. A set of \textit{h} filter vertices (solid green) forward only events in a region-of-interest (ROI) to the \textit{n} particle vertices (purple). There is $n \times n$ connection between particles to share state and weight information. A single particle updates the ROI or the filter vertices, and forwards the average particle state to the external device.}
    \label{fig:architecture}
\end{figure}

Figure~\ref{fig:architecture} shows the connectivity network of SpiNNaker cores. A region-of-interest (ROI) filter forms the first layer of the network, using multiple cores to distribute the processing load. The output of each ROI filter is sent to all particle nodes. Each particle node shares the particle state $<x, y, r>$ and weight $<w>$ with all others at the end of each update cycle in order to normalise the particle weights, and re-sample according to these weights. A single particle sends a packet to update the ROI and also send out the average state to the output of the SpiNNaker.

\section{Demo, System and Preliminary Results}

We propose a live demonstration of these two ``unconventional'' technologies running the proposed algorithm to track a circular target that is moved in front of the camera, and also when the camera itself is moved by hand. The output from the SpiNNaker is displayed on a laptop screen overlaid on the event-stream directly from the camera. The tracking algorithm is consistent, fast, and robust to false detections.

The ATIS sensor is connected to a \textit{Xilinx Zynq} processor through FPGA, with the arm-core connected to the local TCP/IP network. The SpiNNaker is connected to a second \textit{Xilinx Zynq} through FPGA and is also connected to the local TCP/IP network. The YARP robot middle-ware is used to define, establish, and maintain communication between the two \textit{Xilinx Zynq} as well as a single laptop computer that is used for visualisation. The advantage of using YARP is to provide a framework to easily connect the sensors together in a modular way. For example, the CPU version of the algorithm can instead be run, requiring only the connection of the output of the camera to be connected to the CPU module.

Preliminary results indicate that the number of particles can be scaled up to the number of cores available; on a single 48 chip board it is over 500 after the 240  ROI vertices are instantiated. The overhead from scaling the number of particles includes only the communication latency, and is much less than a traditional CPU implementation which must also scale with the algorithm compute time. Typical communication latency is  $< 4.6 \mu s$ per particle. The main drawback of the particle filter algorithm, which applies a limitation on the SpiNNaker platform, is that the particles need to synchronise and wait till all other particles are finished before processing is continued. In comparison, in a fully independent feed-forward network, each vertex always processes despite the state of other vertices. However, we still manage to achieve update rates of over 1 kHz with hundreds of particles, which is not achieved on a standard CPU.

\section*{ACKNOWLEDGEMENT}

This project was instigated at the Capocaccia Cognitive Neuromorphic Engineering Workshop.

\bibliography{bibliography}

\end{document}